%% file: iclr2020_conference.tex
\documentclass{article} % For LaTeX2e
\usepackage{iclr2020_conference,times}

\input{math_commands.tex}

\usepackage{hyperref}
\usepackage{url}
\usepackage{algorithm,algpseudocode} % http://ctan.org/pkg/{algorithms,algorithmicx}
\usepackage{graphicx}
\usepackage{subcaption}
\usepackage{wrapfig}
\usepackage{booktabs}
\usepackage{xcolor}

\graphicspath{ {./images/} }

\title{An Inter-Layer Weight Prediction and Quantization for Deep Neural Networks based on Smoothly Varying Weight Hypothesis}

\author{Kang-Ho Lee, JoonHyun Jeong, and Sung-Ho Bae\thanks{Corresponding author} \\
Department of Computer Science and Engineering\\
Kyung Hee University\\
\texttt{\{ho7719,doublejtoh,shbae\}@khu.ac.kr}
}

\iclrfinalcopy % Uncomment for camera-ready version, but NOT for submission.
\begin{document}

\maketitle

\begin{abstract}
% Network compression for deep neural networks has become an important part
% of deep learning research, because of increased demand for deep learning models
% in practical resource-constrained environments.
Due to a resource-constrained environment, network compression has become an important part of deep neural networks research. In this paper, we propose a new compression method, \textit{Inter-Layer Weight Prediction} (ILWP) and quantization method which quantize the predicted residuals between the weights in all convolution layers based on an inter-frame prediction method in conventional video coding schemes. Furthermore, we found a phenomenon \textit{Smoothly Varying Weight Hypothesis} (SVWH) which is that the weights in adjacent convolution layers share strong similarity in shapes and values, i.e., the weights tend to vary smoothly along with the layers. Based on SVWH, we propose a second ILWP and quantization method which quantize the predicted residuals between the weights in adjacent convolution layers. Since the predicted weight residuals tend to follow Laplace distributions with very low variance, the weight quantization can more effectively be applied, thus producing more zero weights and enhancing the weight compression ratio. In addition, we propose a new \textit{inter-layer loss} for eliminating non-texture bits, which enabled us to more effectively store only texture bits. That is, the proposed loss regularizes the weights such that the collocated weights between the adjacent two layers have the same values. Finally, we propose an ILWP with an inter-layer loss and quantization method. Our comprehensive experiments show that the proposed method achieves a much higher weight compression rate at the same accuracy level compared with the previous quantization-based compression methods in deep neural networks.
\end{abstract}

\section{Introduction}
Deep neural networks have demonstrated great performance for various tasks in many fields,
such as image classification
(\citealt{lecun1990handwritten}; \citealt{krizhevsky2012imagenet}; \citealt{he2016deep}),
object detection
(\citealt{ren2015faster}; \citealt{he2017mask}; \citealt{redmon2018yolov3}),
image captioning \citep{jia2015guiding}, and speech recognition
(\citealt{hinton2012deep}; \citealt{xiong2018microsoft}).
Wide and deep neural networks achieved great accuracy with the aid of
the enormous number of weight parameters and high computational cost
(\citealt{simonyan2014very}; \citealt{he2016deep}; \citealt{huang2017densely}).

However, as demands toward constructing the neural networks in the resource-constrained
environments have been increasing, making the resource-efficient neural network
while maintaining its accuracy becomes an important research area of deep neural networks.
Several studies have aimed to solve this problem.

In \citet{lecun1990optimal}, \citet{hassibi1993second}, \citet{han2015learning}
and \citet{li2016pruning}, network pruning methods were proposed for memory-efficient
architecture, where unimportant weights were forced to have zero values in terms of accuracy.
In \citet{fiesler1990weight}, \citet{gong2014compressing} and \citet{han2015deep},
weights were quantized and stored in memory, enabling less memory usage of deep neural networks.
On the other hand, some literature decomposed convolution operations into sub operations
(e.g., depth-wise separable convolution) requiring less computation costs
at similar accuracy levels (\citealt{howard2017mobilenets}; \citealt{zhang2018shufflenet};
\citealt{sandler2018mobilenetv2}; \citealt{ma2018shufflenet}).

In this paper, we show that the weights between the adjacent two convolution layers
tend to share high similarity in shapes and values.
We call this phenomenon \textit{Smoothly Varying Weight Hypothesis} (SVWH).
This paper explores an efficient neural network method
that fully takes the advantage of SVWH.

Specifically, inspired by the prediction techniques widely used in
video compression field (\citealt{wiegand2003overview}; \citealt{sullivan2012overview}),
we propose a new weight compression scheme based on an inter-layer weight prediction 
technique, which can be successfully incorporated into
the depth-wise separable convolutional blocks
(\citealt{howard2017mobilenets}; \citealt{zhang2018shufflenet};
\citealt{sandler2018mobilenetv2}; \citealt{ma2018shufflenet}). 

\textbf{Contributions:} Main contributions of this paper are listed below:
\begin{itemize}
    \item From comprehensive experiments, we find out that the weights between
    the adjacent layers tend to share strong similarities,
    which lead us to establishing SVWH. 
    \item Based on SVWH, we propose a simple and effective
    \textit{Inter-Layer Weight Prediction} (ILWP) and quantization framework
    enabling a more compressed neural networks than only applying quantization
    on the weights of the neural networks.
    \item To further enhance the effectiveness of the proposed ILWP,
    we devise a new regularization function, denoted as \textit{inter-layer loss},
    that aims to minimize the difference
    between collocated weight values in the adjacent layers,
    resulting in significant bit saving for non-texture bits
    (i.e., bits for indices of prediction).   
    \item Our comprehensive experiments demonstrate that, the proposed scheme
    achieves about 53\%  compression ratio on average in 8-bit quantization
    at the same accuracy level compared to the traditional quantization method
    (without prediction) in both MobileNetV1 \citep{howard2017mobilenets} and
    MobileNetV2 \citep{sandler2018mobilenetv2}.
\end{itemize}

\section{Related Work}
\label{related_work}

\textbf{Network pruning:} Network pruning methods prune the unimportant weight parameters,
enabling to reduce the redundancy of weight parameters inherent in neural networks.
\citet{lecun1990optimal} and \citet{hassibi1993second} reduced the number of weight
connections implicitly through setting an proper objective function for training.
\citet{han2015learning} successfully removed the unimportant weight connections
through certain thresholds for the weight values, showing no harm of accuracy
in the state-of-the-art convolutional neural network models.
Recently, structured (filter/channel/layer-wise) pruning methods have been proposed
in \citet{liu2017learning} and \citet{li2016pruning}, where a set of weights is pruned
based on certain criteria (e.g., the sum of absolute values in the set of weights),
demonstrating significantly reduced number of weight parameters and computational costs.
Furthermore, \citet{he2018amc} use AutoML for channel pruning and their proposed method
get accuracy 13.2\% more than filter pruning method \citep{li2016pruning}.
Our paper is linked to the pruning methods in perspective of assigning more zero weights
for weight compression.

\textbf{Quantization:} Quantization reduces the representation bits of original weights
in neural networks. \citet{fiesler1990weight} proposed a weight quantization using
weight discretization in neural networks. \citet{han2015learning} incorporated
a vector quantization into the pruning, proving that quantization and pruning
can jointly work for weight compression without accuracy degradation.
This pruning-quantization framework, i.e., called Deep Compression, became a milestone
in model compression research of deep neural networks. \citet{lin2016fixed} proposed
a fixed-point quantization using a linear scale factor for weight values
where bit-widths for quantization are adaptively found for each layer,
thus enabling 20\% reduction of the weight size in memory without any loss
in accuracy compared to the baseline fixed-point quantization method.
Furthermore, \citet{sung2015resiliency}, \citet{zhuang2018towards} and
\citet{zhao2019improving} use clipping weights before applying linear quantization and
those methods are improve accuracy than linear quantization without clipping.

For 1-bit quantization which is named as Binary Neural Networks, several studies
(\citealt{BinaryConnect}; \citealt{BNN}; \citealt{QNN}) binarized the weights and/or
activations in the process of back-propagation, enjoying considerably reduced usage of
memory space and computation overhead.

Compared to the aforementioned quantization techniques, our work applies quantization
in the combination of the residuals for the weights in inter layers rather than
the weight values themselves. Empirically, we found that the residuals tend to
produce much larger portion of zero values when quantized, since they often
follow very narrow Laplace distributions. Therefore, our proposed method can
significantly reduce the memory consumption for neural networks as shown
in Section \ref{experiments}.

\textbf{Prediction in conventional video coding:} Prediction technique is considered one of the most crucial parts in video compression, aiming at minimizing the magnitudes of signals to be encoded by subtracting the input signals to the most similar encoded signals in a set of prediction candidates (\citealt{wiegand2003overview}; \citealt{rijkse1996h}; \citealt{sullivan2012overview}).
The prediction methods can produce the residuals of signals with low magnitudes as well as a large number of non/near zero signals. Therefore,  they have effectively been incorporated into transforms and quantization for concentrating powers in low frequency regions and reducing the entropy, respectively. There are two prediction techniques: inter- and intra- predictions. The inter-prediction searches the best prediction signals from the encoded neighbor frames out of the current frame, while the intra-prediction generates a set of prediction signals from the input signals and determine the best prediction (\citealt{wiegand2003overview}; \citealt{sullivan2012overview}). This is because intra frames that use only intra-prediction for compression are used as reference frames for subsequent frames to be predicted. 

Note that a few studies explored to apply the transform techniques of video and/or image coding to the weight compression problem in neural networks. \citet{wang2016cnnpack} and \citet{ko2017adaptive} applied DCT (Discrete Cosine Transform) used in the JPEG (Joint Picture Encoding Group) algorithm to the model compression problem of deep neural networks such that the energy of weight values became concentrated in low frequency regions, thus producing more non/near-zero DCT coefficients for the weights. Compared to the aforementioned papers, our work does not adopt transform techniques to reduce model sizes, because the transforms introduce much computation in inference, decreasing the effectiveness of the weight compression in practical applications.

In this paper, we found out that the inter-prediction technique can play
a crucial role for weight compression under the SVWH condition
(i.e., the weights between adjacent layers tend to be similar).
The proposed \textit{inter-layer loss} reinforces SVWH for reducing the non-texture bits
in the training process. As a result, the proposed inter-prediction and quantization
framework for weight compression yields impressive compression performance enhancement
at the similar accuracy level compared to the baseline models.

\section{Method}
\label{method}

\subsection{Basic Inter-Layer Weight Prediction}
\label{basic_interlayer_weight_prediction}

\begin{figure}[!t]
\centering
    \includegraphics[scale=0.185]{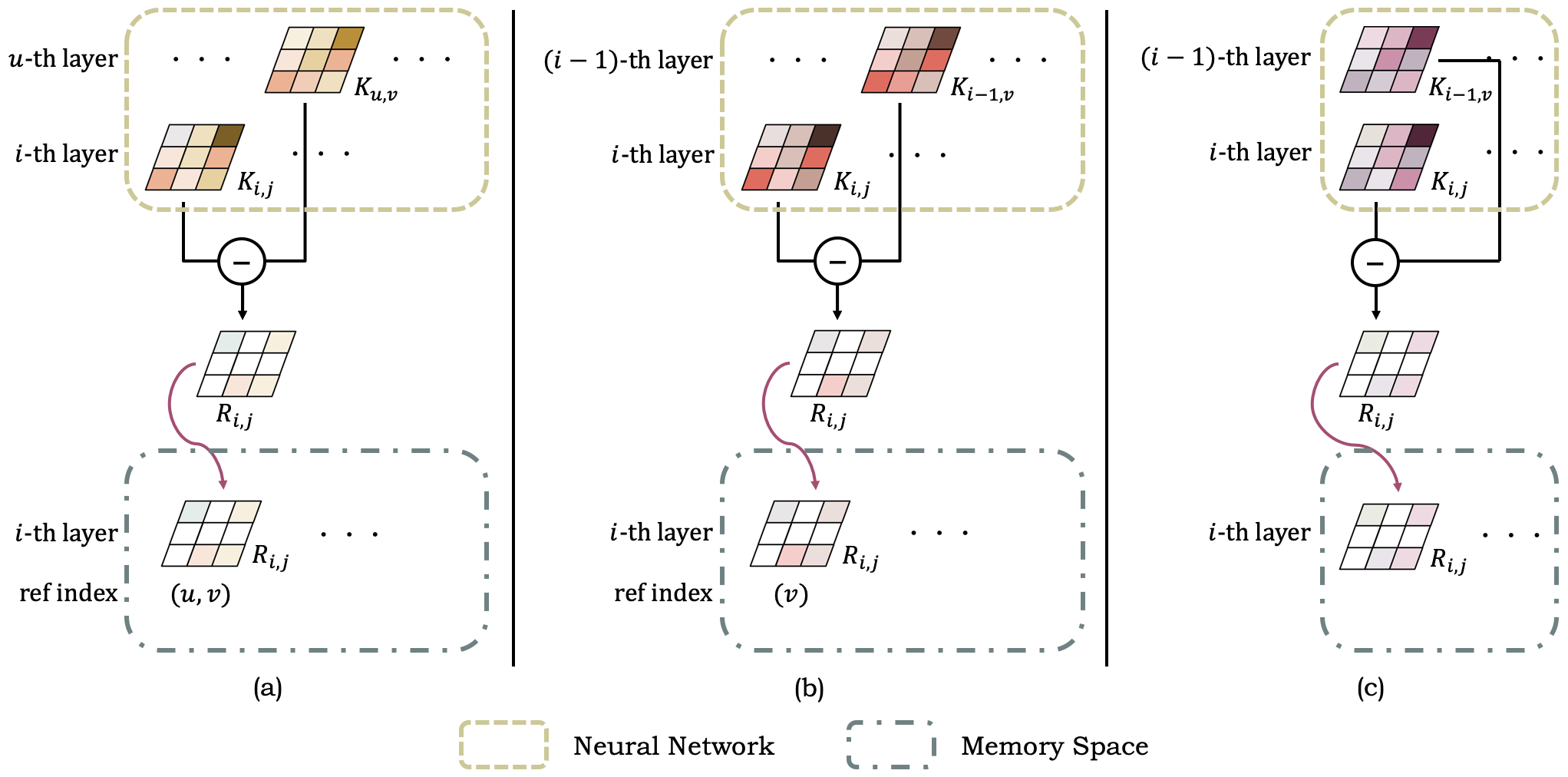}
    \centering
\caption{Our \textit{Inter-Layer Weight Prediction} (ILWP) scheme.
(a) ILWP with the full search strategy described in
Section \ref{basic_interlayer_weight_prediction};
(b) ILWP with the local search strategy described in
Section \ref{smoothly_varying_weight_hypothesis};
(c) ILWP trained with inter-layer loss and without searching described in
Section \ref{inter-layer_loss}.
The ref index is the referenced kernel index for ILWP, i.e. the index of
the most similar weight kernel on target layer. Other notation is described
in each subsection.}
% \caption{Our Inter-Layer Weight Prediction scheme of (1) finding the most similar kernel $K_{u,v}$ at all previous layer ($u<i$), using $L_{1}$ distance; (2) calculating the residuals between two kernels; (3) saving the residuals with the index of the reference kernel, i.e., the best prediction, into memory space.}
\label{fig:inter-layer_weight_prediction}
\end{figure}

Figure \ref{fig:inter-layer_weight_prediction}-(a) shows an simple example
of the proposed ILWP method with a full search strategy (ILWP-FSS).
The full search strategy (FSS) indicates that, for the $j$-th weight in the kernel
of the $i$-th convolution layer ($K_{i,j}$), it searches the most similar weight kernel
(i.e., $K_{u,v}$ in Figure \ref{fig:inter-layer_weight_prediction}-(a))
in the range of [1, $i-1$]-th layers given a pre-trained neural network model.
We then compute the residuals ($R_{i,j}$) between the current weight and
the best prediction, which is finally quantized in a certain bit representation and
stored in memory space with the index of the best prediction (i.e., $(u,v)$
in Figure \ref{fig:inter-layer_weight_prediction}-(a)).

The proposed ILWP is performed from the second layer to the last layer in a sequential
manner. It should be noted that, because of the large portion of non-texture bits
(indices of the best prediction), ILWP-FSS tends to produce more bits
(both texture (residual) and non-texture bits) than the standard weight values
without ILWP, which makes ILWP worthless in compressing the weights. In the next
subsection, we describe how SVWH solves this problem effectively.

\subsection{Smoothly Varying Weight Hypothesis}
\label{smoothly_varying_weight_hypothesis}

\begin{figure}[!t]
\centering
    \includegraphics[scale=0.35]{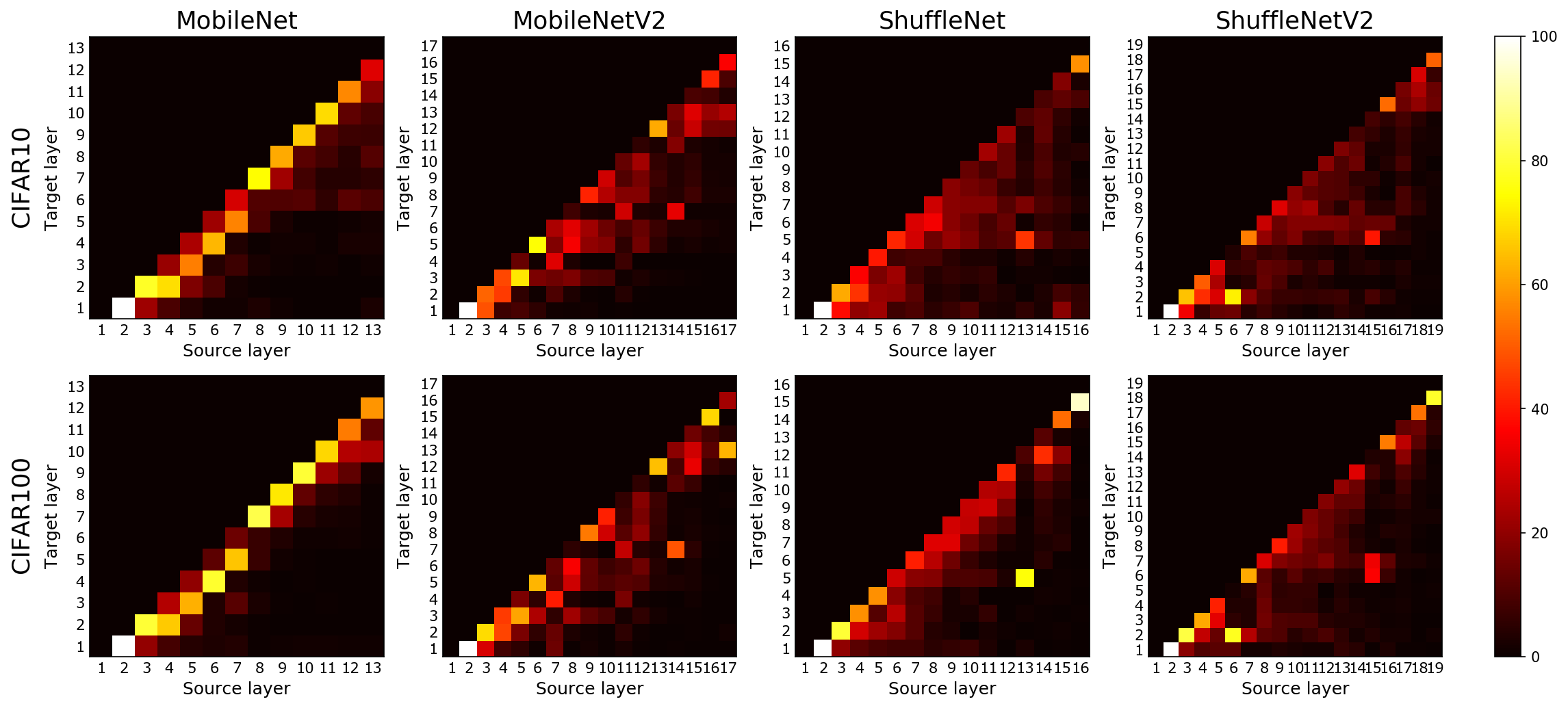}
    \centering
\caption{Heatmaps that show the percentage in the number of best predictions
with respect to minimizing $L_{1}$ distance between the source layer ($x$-axis)
and the target layer ($y$-axis) in the neural networks trained on CIFAR10 (top)
and CIFAR100 (bottom). From left to right are MobileNet, MobileNetV2, ShuffleNet,
and ShuffleNetV2}
\label{fig:heatmap_baseline}
\end{figure}

As shown in Figure \ref{fig:heatmap_baseline}, the dominant portions of the best predictions in the current ($i$-th layer) layer are obtained from its previous ($(i-1)$-th layer) layers, being consistently observed in all the layers of the neural networks. From these observations, we claim that the adjacent layers tend to consists of the similar weight values, leading us to propose SVWH. The proposed SVWH can be mathematically expressed as 

\begin{equation}
\mathrm{P}[L(K_{i,j},K_{i-1,v}) >  L(K_{i,j},K_{u,v})] \: > \: \mathrm{P}[L(K_{i,j},K_{i-1,v}) \leq L(K_{i,j},K_{u,v})],
\label{eq:svwh}
\end{equation}

where $\mathrm{P}[\cdot]$ is a probability, $L(\cdot,\cdot)$ is a distance between two kernels, and $(u, v)$ are the indices of the layer and kernel, where  $1 \leq u < i-1$ (See Figure \ref{fig:inter-layer_weight_prediction}).

Inspired by SVWH, we propose an enhanced version of ILWP, i.e. ILWP
with a local search strategy (ILWP-LSS) that finds the best prediction only
from the previous layer (See Figure \ref{fig:inter-layer_weight_prediction}-(b)).
The local search strategy (LSS) can effectively reduce the non-texture bits,
because the non-texture bits for the layer index are not required.
Moreover, the LSS allows the network to keep the weights only
in the previous one layer in inference, thus reducing the memory buffer for
keeping the weights. Our experimental results in Section \ref{experiments} show that,
the local search strategy enhances the compression performance compared to
the FSS in the ILWP framework of deep neural networks.
In this paper, to further increase the effectiveness of the proposed ILWP, we devise
a new regularization loss, namely inter-layer loss.

\subsection{ILWP with Inter-Layer Loss}
\label{inter-layer_loss}

To further exploit SVWH, we propose a new inter-layer loss which makes the collocated
filter weights between the adjacent layers have almost the same values. We find out
that our ILWP can more effectively be applied to the depth-wise convolution
($3\times3$ spatial convolution) layer in the depth-wise separable convolution block
\citep{howard2017mobilenets}, compared to the traditional 3D convolutions
($3\times 3 \times C$ convolution). This is because high dimensionality of the weights
in the traditional 3D convolution filter tends to hinder finding out the best prediction
sharing strong similarity. This can introduce a longer tail and wider shape of
the Laplace distribution for the residuals of the weights, thus decreasing
the compression efficiency. Moreover, it is not possible to predict the weight kernels
having different channel dimensions from the current weight kernel. This can limit
the usability of the proposed ILWP.

On the other hand, the depth-wise convolution consists only of nine elements,
and all the depth-wise convolutions have canonical kernel shapes with $3\times3$ size
in whole networks. Also, since the point-wise convolution ($1\times 1$ convolution)
learns channel correlations for the output features of the depth-wise convolution layer,
forcing the collocated weights to be similar does hardly affect the accuracy of
neural networks. These characteristics of depth-wise convolution enhance
the usability of the proposed ILWP in the weight compression of neural networks.

Moreover, it becomes more popular to use depth-wise separable convolutions in the recent neural network architectures, due to its high efficiency (\citealt{kaiser2017depthwise}; \citealt{sandler2018mobilenetv2}; \citealt{zhang2018shufflenet}; \citealt{ma2018shufflenet}). Therefore, we apply the proposed method into the spatial convolution in the depth-wise separable convolution block. Our proposed \textit{inter-layer loss} can dramatically eliminate the non-texture bits and is defined as follows:

\begin{equation}
\mathcal{L}_{\text{inter-layer}} = \frac {1} {Z} \sum_{i=2}^{N} \sum_{j=1}^{c_{i-1}} \left\vert K_{i,j} - K_{i-1,v} \right\vert,
\label{eq:inter-layer-loss}
\end{equation}

where $Z$ is the number of weights to be predicted, and $N$ is the number of depth-wise convolution layers in a whole neural network. In Eq. (\ref{eq:inter-layer-loss}), $v$ is the index in the previous layer which is equal to $j \bmod c_{i-1}$ in the inter layer loss. 
The proposed loss function in Eq. (\ref{eq:inter-layer-loss}) not only regularizes the weights but also allows us to eliminate all the non-texture bits for indices of the best predictions, since the network can always predict the weight values of the current layer from the collocated weights in its previous layer. For training the neural networks with our proposed loss, our total loss is formulated as follows:

\begin{equation}
\mathcal{L}_{\text{total}} = \mathcal{L}_{\text{cls}} + \lambda \mathcal{L}_{\text{inter-layer}},
\label{eq:total-loss}
\end{equation}

where $\mathcal{L}_{\text{cls}}$ is the conventional classification loss using
the cross-entropy loss \citep{he2016deep}. $\lambda$ in Eq. (\ref{eq:total-loss}) is
the control parameter for the \textit{inter-layer loss} and is set to 1
over all experiments. From our comprehensive experiments, we found that setting
$\lambda$ in Eq. (\ref{eq:total-loss}) to 1 is suitable to match the trade-off between
the performance of neural networks and the parameter size of neural networks
(See Figure \ref{fig:lambda_study} in Section \ref{optimal_lambda}).

\begin{figure}[!t]
\centering
    \includegraphics[scale=0.35]{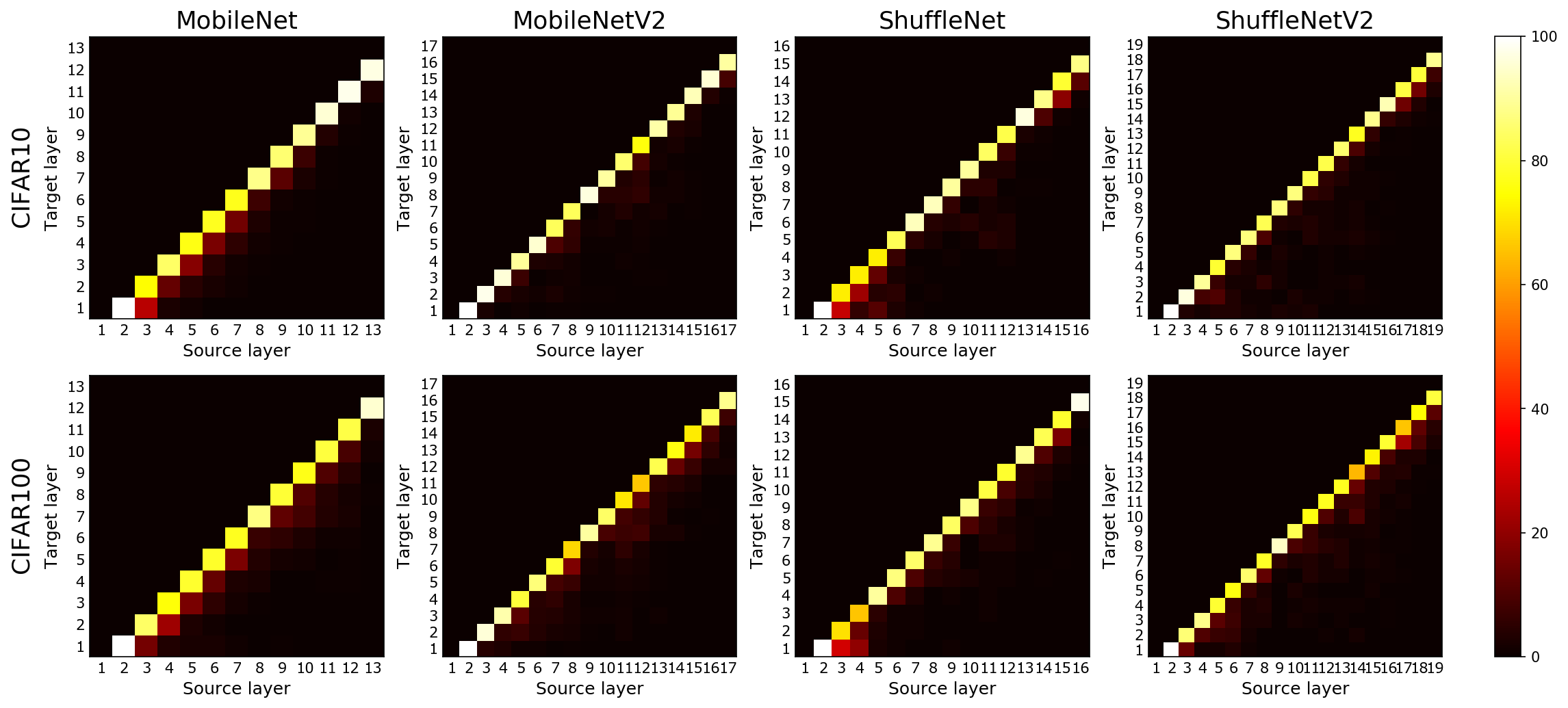}
    \centering
\caption{Heatmaps that show the percentage in the number of best predictions
with respect to minimizing $L_{1}$ distance between the source layer ($x$-axis)
and the target layer ($y$-axis) in the neural networks trained on CIFAR10 (top)
and CIFAR100 (bottom) with \textit{inter-layer loss}. From left to right are MobileNet,
MobileNetV2, ShuffleNet, and ShuffleNetV2}
\label{fig:heatmap_interlayer_loss}
\end{figure}

Through our new \textit{inter-layer loss}, SVWH is explicitly controlled in
a more elaborate manner as shown in Figure \ref{fig:heatmap_interlayer_loss}
that shows the Heatmaps for the percentage in the number of best predictions
with respect to minimizing $L_1$ distance between the source layer ($x$-axis)
and the target layer ($y$-axis) in MobileNet, MobileNetV2, ShuffleNet, and ShuffleNetV2 trained with
the proposed inter-layer loss. Finally, the reconstruction of the current weight kernel
is performed as follow:

\begin{equation}
\hat{K}_{i,j} = \tilde{R}_{i,j} + \hat{K}_{i-1,v},
\label{eq:calc-residual-inter-layer-loss}
\end{equation}

where $\tilde{R}_{i,j}$ is the quantized residual at the $i$-th layer and $j$-th
filter position. $\hat{K}$ in Eq. (3) is the final reconstruction of $K$. Note that,
to reconstruct the current weights, the weights only in the previous layer
(i.e., $\hat{K}_{i-1,v}$ in Eq. (\ref{eq:calc-residual-inter-layer-loss}))
are required (See Figure \ref{fig:inter-layer_weight_prediction}-(c)).
The residuals of the weights are quantized through a linear quantization technique,
and then saved using Huffman coding for the purpose of efficient model capacity
\citep{han2015learning}. Thanks to the prediction scheme, our method usually remains
more high-valued non-zero weights than the traditional quantization methods.
Since high weight values importantly contribute to the prediction performance
\citep{han2015learning}, our method can retain both high accuracy and
weight compression performance.

The weight kernels in the first layer are not quantized since it is as important as Intra-frames (i.e. reference frames) in a group of pictures of the video coding scheme \citep{wiegand2003overview}. If the weight kernels in the first layer are quantized, the weight kernels in subsequent layers which are predicted based on the weight kernels in the first layer are negatively affected, leading to accuracy drop in neural networks.

\section{Experiments}
\label{experiments}

In this section, we describe and prove the superiority of the proposed method by applying
it on image classification tasks, specifically for CIFAR-10 and CIFAR-100 datasets.
For securing generality of our proposed IWLP, we applied our method in
four convolutional neural networks, MobileNet, MobileNetV2, ShuffleNet, and ShuffleNetV2.
Four NVidia 2080-Ti GPU with the Intel i9-7900X CPU are used to perform the experiments.
For the hyper-parameter setting in the training process, we set the initial learning rate
as 0.1, which is multiplied by 0.98 every epochs.
We used Stochastic Gradient Descent (SGD) optimizer with Nesterov momentum
\citep{sutskever2013importance} factor 0.9. All the neural networks are trained
for 200 epochs with a batch size of 256.
The baseline model is each aforementioned four convolutional neural networks
which is quantized by linear quantization on the weights of depthwise convolutional kernel.
In all the experiments, the test accuracy and parameters in kilobyte (KB)
are marked from the average of 5 runs.

\subsection{Optimal $\lambda$ selection}
\label{optimal_lambda}

\begin{figure}[!t]
\centering
    \includegraphics[scale=0.44]{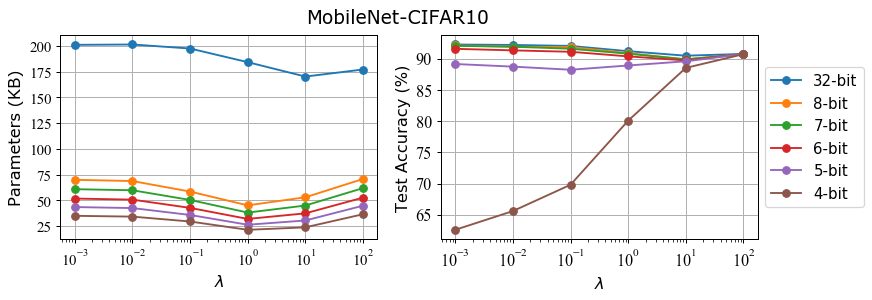}
    \centering
\caption{Comparison of parameter size in kilobytes (KB) and top-1 test accuracy in MobileNet trained on CIFAR-10 dataset for different $\lambda$ in Eq. (\ref{eq:total-loss}) after $n$-bit linear quantization. 32-bit means using full precision floating point.}
\label{fig:lambda_study}
\end{figure}

First, we found the optimal $\lambda$ in Eq. (\ref{eq:total-loss}) that is suitable to match the trade-off between the accuracy performance and the parameter size of neural networks. Figure \ref{fig:lambda_study} shows the parameter sizes and top-1 test accuracy in MobileNet trained on the CIFAR-10 dataset for different $\lambda$ in Eq. (\ref{eq:total-loss}) after quantization and Huffman coded. We can see that the case of $\lambda = 1$ has the smallest parameter size in neural networks, and slightly lower top-1 test accuracy in the CIFAR-10 dataset. Furthermore, not only MobileNet trained on CIFAR-100, but also other models, i.e., MobileNetV2, ShuffleNet, and ShuffleNetV2 trained on CIFAR-10 and CIFAR-100 datasets also has very similar results. So, we experimented by setting the $\lambda$ in Eq. (\ref{eq:total-loss}) to 1.

\subsection{Experimental results}
\label{experimental_results}

\begin{figure}[!t]
\centering
    \includegraphics[scale=0.36]{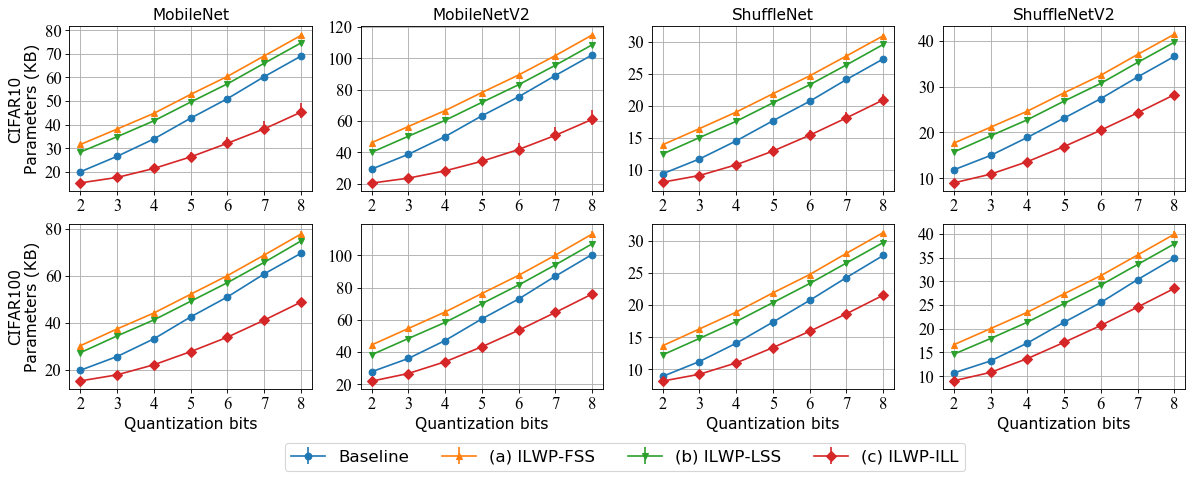}
    \centering
\caption{Comparison of Parameter size in kilobytes (KB) for different quantization bits in $\{2, 3, 4, ..., 8\}$ on CIFAR-10 (top) and CIFAR-100 (bottom) datasets. From left to right are MobileNet, MobileNetV2, ShuffleNet, and ShuffleNetV2. (a) ILWP with the full search strategy (ILWP-FSS); (b) ILWP with the local search strategy (ILWP-LSS); (c) ILWP trained with inter-layer loss and without searching (ILWP-ILL) described in Section \ref{inter-layer_loss}. In all the figures, each dot represents the average test accuracy over 5 training times in each quantization bit in $ \{ 2, 3, 4, ..., 8 \} $.}
\label{fig:acc_param}
\end{figure}

Figure \ref{fig:acc_param} shows the parameter size of our proposed methods for different
quantization bits in $\{2, 3, 4, ..., 8\}$ on CIFAR-10 and CIFAR-100 datasets.
Figure \ref{fig:acc_param} shows that our proposed ILWP-ILL yields higher compression ratio in the weight parameters. On the other hand, ILWP-FSS and ILWP-LSS have larger parameter sizes than the baseline. This is because both ILWP-FSS and ILWP-LSS contain non-texture bits (bits for indices of the best predictions), resulting in bit overheads.

%Baseline에 대한 설명 추가

For further analysis on the texture and non-texture bits in the proposed ILWP, Table \ref{table:parameter_size} shows the performance comparison of our proposed methods in terms of the parameter sizes for the depth-wise convolution kernels and top-1 test accuracy. Due to the inter-layer prediction scheme, all the proposed methods show less amount of texture bits compared to the baseline. However, the total amounts of bits of ILWP-FSS and ILWP-LSS are higher than the baseline model, which is due to the presence of the non-texture bits ($u$ and $v$ in Figure \ref{fig:inter-layer_weight_prediction}). However, ILWP-ILL demonstrated much reduced amount of total bits compared to the baseline, which is due to the property that this method does not require saving any indices of reference kernels while maintaining good accuracy under the SVWH condition.

\begin{table}[!t]
\caption{Parameter size (texture bits, non-texture bits, total bits) in kilobytes (KB) and top-1 test accuracy of depth-wise convolution kernels after 8 bit quantization and Huffman coding, in MobileNet trained on CIFAR-10. We compared the performance of the three variations of ILWP: ILWP-FSS, ILWP-LSS, and ILWP-ILL.}
\label{table:parameter_size}
\centering
\begin{tabular}{@{~~}l|@{~~~}c@{~~~~}c@{~~~}|@{~~~}c@{~~~}|@{~~~}c@{~~}}
    \toprule
        \multicolumn{1}{c|@{~~~}}{\textbf{Method}} & \textbf{texture bits} & \textbf{non-texture bits} & \textbf{total bits} & \textbf{Test accuracy}\\
    \midrule
        Baseline & 70.192 & - & 70.192 & 92.125\\
    \midrule
        ILWP-FSS & 67.716 & 11.177 & 78.893 & 92.225\\
        ILWP-LSS & 67.415 & 8.540 & 75.955 & \textbf{92.235}\\
        ILWP-ILL & \textbf{33.373} & - & \textbf{33.373} & 91.48\\
    \bottomrule
\end{tabular}
\end{table}

\begin{figure}[!t]
\centering
    \includegraphics[scale=0.36]{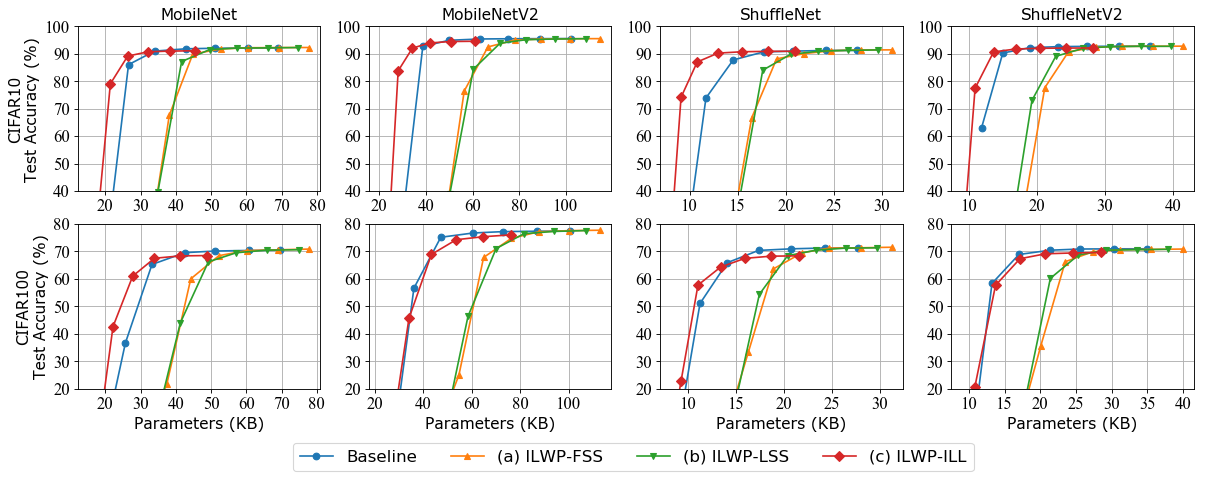}
    \centering
\caption{Performance results of our proposed methods on CIFAR-10 (top) and CIFAR-100 (bottom) datasets for the parameter size in kilobytes (KB) and test accuracy. From left to right are MobileNet, MobileNetV2, ShuffleNet, and ShuffleNetV2. $x$-axis: the total size of the quantized and Huffman coded depth-wise convolutional kernels in storage; $y$-axis: Top-1 test accuracy. (a) ILWP-FSS; (b) ILWP-LSS; (c) ILWP-ILL. In all the figures, each dot represents the average test accuracy over 5 training times in each quantization bit in $ \{ 2, 3, 4, ..., 8 \} $.}
\label{fig:ablation_study}
\end{figure}

Figure \ref{fig:ablation_study} shows the results of our proposed methods on CIFAR-10 and CIFAR-100 datasets for the parameter size in kilobytes (KB) and test accuracy. As shown in Figure \ref{fig:ablation_study}, the ILWP-ILL outperforms the other compared ones in terms of the trade-off between the amounts of compressed bits and accuracy. This is because ILWP-ILL significantly saves the weight bits by eliminating the non-texture bits as well as increasing the effectiveness of the quantization process as the residuals tend to follow much narrower Laplace distributions than the original weight values (See Section \ref{analysis}).

Compared to ILWP-FSS, ILWP-LSS shows very slight improvement in trade-off between the size of parameters and accuracy as ILWP-LSS does not store the layer index of the best prediction ($u$ in Figure 1). This is due to the fact that the portion of the bits for the layer indices is much smaller than the portion of the bits for the kernel indices. Compared to the baseline models, it is worth noting that both ILWP-FSS and ILWP-LSS have often worse compression efficiency compared to the baseline. This is because they introduce non-texture bits consisting of a large portion in the total bits for weight parameters. Therefore, it can be concluded that our foundation of SVWH and the proposed inter-layer loss allow the network to make use of full advantages in the inter-layer prediction scheme of the conventional video coding frameworks.

\section{Analysis}
\label{analysis}

\begin{figure}[ht]
\centering
\begin{subfigure}{0.325\textwidth}
    \includegraphics[scale=0.45]{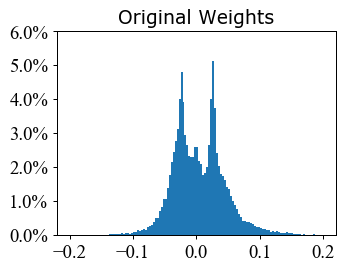}
    \centering
    \caption{}
\end{subfigure}
\begin{subfigure}{0.325\textwidth}
    \includegraphics[scale=0.45]{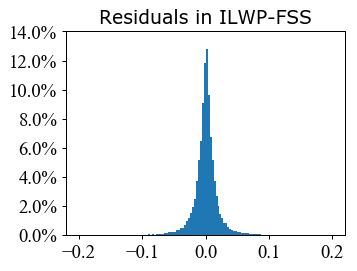}
    \centering
    \caption{}
\end{subfigure}
\begin{subfigure}{0.325\textwidth}
    \includegraphics[scale=0.45]{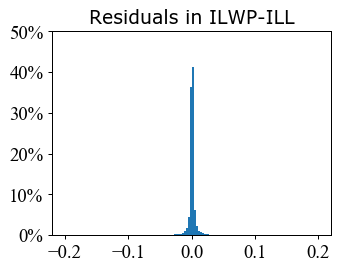}
    \centering
    \caption{}
\end{subfigure}
\caption{Comparison of the distributions for the weights and residuals in all the depth-wise convolution kernels in MobileNet trained on CIFAR-100. (a) Distribution of the weight kernels in baseline model; (b) Distribution of residuals in ILWP-FSS; (c) Distribution of the residuals in ILWP-ILL.}
\label{fig:comparison_distribution}
\end{figure}

Figure \ref{fig:comparison_distribution} compares the distributions for the weights and
residuals in all the depth-wise convolution kernels in MobileNet trained on CIFAR-100.
Figure \ref{fig:comparison_distribution}-(a), -(b), and -(c) visualize the distribution
of the weight kernels in baseline model, the distribution of residuals in ILWP-FSS and
the distribution of the residuals in ILWP-ILL, respectively. As shown in Figure
\ref{fig:comparison_distribution}, ILWP-FSS produces a single modal Laplace distribution.
This nice property contributes to high compression capacity for storing the weights
in neural networks when these residuals are quantized and Huffman coded.
However, this method requires saving the additional indices, leading to
consuming more amount of bits.

As shown in Figure \ref{fig:comparison_distribution}-(c), it can be observed that
for ILWP-ILL, the residuals are located in near-zero positions (about 46\% of the weights),
following a very sharp Laplace distribution. This indicates that ILWP-ILL with quantization
allows neural networks to achieve remarkable weight compression performance by generating
a large amount of zero coefficients after quantization. Furthermore, in terms of
information theory, the Shannon entropy $H_{x}$ of Laplace distribution is derived
as follows\footnote{See Section 2.1.3.5 (p.21) in \citet{kotz2012laplace}}:

\begin{equation}
\begin{aligned}
H_{x} & = -\int_{-\infty}^{\infty}f_{L}(x)\log(f_{L}(x))\\
& = -\int_{-\infty}^{\infty}(\frac{1}{2b}e^{-\frac{|x-\mu|}{b}})\log(\frac{1}{2b}e^{-\frac{|x-\mu|}{b}})\\
& = \log(2b) + 1,
\end{aligned}
\label{eq:entropy_ILL}
\end{equation}

where, $f_{L}(\cdot)$ is the probability density function of the Laplace distribution and $b$ and $\mu$ is scale and location factors of the Laplace distribution, respectively. So, The Shannon entropy is proportional to scale parameter $b$, which controls width of Laplace distribution, i.e., small $b$ has a narrow Laplace distribution.

As shown in Figure \ref{fig:comparison_distribution}, it is observed that distribution of residuals in ILWP-ILL has much narrower Laplace distribution than distribution of residuals in ILWP-FSS. Consequently, the information entropy of the distribution of the residuals in ILWP-ILL is lower than the information entropy of the distribution of the residuals in ILWP-FSS. Meanwhile, the entropy coding as Huffman coding is more compressed in small information entropy. As a result, the ILWP-ILL method is more compressed than the ILWP-FSS method after quantization and entropy coding.

\section{Conclusion}
\label{conclusion}
We propose a new inter-layer weight prediction with inter-layer loss for efficient deep neural networks. Motivated by our observation that the weights in the adjacent layers tend to vary smoothly, we successfully build a new weight compression framework combining the inter-layer weight prediction scheme, the inter-layer loss, quantization and Huffman coding under SVWH condition.
Intuitively, our prediction scheme significantly decreases the entropy of the weights by making them much narrower Laplace distributions, thus leading remarkable compression ratio of the weight parameters in neural networks. Also, the proposed inter-layer loss effectively eliminates the non-texture bits for the best predictions.  To the best of our knowledge, this work is the first to report the phenomenon of the weight similarities between the neighbor layers and to build a prediction-based weight compression scheme in modern deep neural network architectures.

% 1st contribution 
% 2nd contribution Inter-Layer Loss
% 3rd contribution the first study of predicting the weights in deep neural networks.


\bibliography{iclr2020_conference}
\bibliographystyle{iclr2020_conference}

% \appendix
% \section{Optimal $\lambda$ Search in other models in CIFAR-10/CIFAR-100 datasets}
% \label{appendix_lambda_search}

\end{document}

%% file: math_commands.tex
\usepackage{amsmath,amsfonts,bm}

\def\eqref#1{equation~\ref{#1}}
\def\1{\bm{1}}

\DeclareMathAlphabet{\mathsfit}{\encodingdefault}{\sfdefault}{m}{sl}
\SetMathAlphabet{\mathsfit}{bold}{\encodingdefault}{\sfdefault}{bx}{n}